\documentclass[pmlr]{jmlr}

\usepackage{longtable}

\usepackage{printlen}
\usepackage{setspace}

\usepackage{microtype}
\usepackage{graphicx}
\usepackage{todonotes}
\usepackage{afterpage}
\usepackage{balance}
\usepackage{multicol}
\usepackage{xcolor}
\usepackage{amsmath}
\usepackage{amssymb}

\usepackage{tikz,pgfplots,pgfplotstable,pgf-pie}
\usepackage{filecontents}
\usepackage{xcolor} 

\usepackage[nameinlink,noabbrev]{cleveref}
\crefname{equation}{}{}
\Crefname{equation}{}{}

\usepackage{siunitx,tikz}

\usepackage{nicefrac}

\usepackage{booktabs}
 \usepackage{siunitx}
 \usepackage{listings}
 \definecolor{codegreen}{rgb}{0,0.6,0}
\definecolor{codegray}{rgb}{0.5,0.5,0.5}
\definecolor{codepurple}{rgb}{0.58,0,0.82}
\definecolor{backcolour}{rgb}{0.95,0.95,0.92}

\lstdefinestyle{mystyle}{
    backgroundcolor=\color{backcolour},   
    commentstyle=\color{codegreen},
    keywordstyle=\color{magenta},
    numberstyle=\tiny\color{codegray},
    stringstyle=\color{codepurple},
    basicstyle=\ttfamily\footnotesize,
    breakatwhitespace=false,         
    breaklines=true,                 
    captionpos=b,                    
    keepspaces=true,                 
    numbers=left,                    
    numbersep=5pt,                  
    showspaces=false,                
    showstringspaces=false,
    showtabs=false,                  
    tabsize=2
}
\theorembodyfont{\upshape}
\theoremheaderfont{\scshape}
\theorempostheader{:}
\theoremsep{\newline}

\jmlrvolume{}
\jmlryear{}
\jmlrworkshop{NeurIPS 2021 Competition and Demonstration Track}

\title[Real Robot Challenge]{Real Robot Challenge: A Robotics Competition in the Cloud \titletag{\thanks{https://real-robot-challenge.com}}}

 \author{\doublespacing
  \Name{Stefan Bauer$^{*}$$^{1}$} 
 \Email{baue@kth.se}\\
 \Name{Manuel W\"uthrich$^{*}$$^{2}$} 
 \Email{manuel.wuthrich@pm.me}\\
   \Name{Felix Widmaier$^{*}$$^{2}$} \Email{felix.widmaier@tuebingen.mpg.de}\\
  \Name{Annika Buchholz$^{2}$} \Email{annika.buchholz@tuebingen.mpg.de}\\
  \Name{Sebastian Stark$^{2}$} \Email{stark@tuebingen.mpg.de}\\
   \Name{Anirudh Goyal$^{3}$} \Email{anirudhgoyal9119@gmail.com}\\
   \Name{Thomas Steinbrenner$^{2}$} \Email{thomas.steinbrenner@tuebingen.mpg.de}\\
  \Name{Joel Akpo$^{2}$} \Email{joel.akpo@is.mpg.de}\\
   \Name{Shruti Joshi$^{3}$} \Email{shruti.joshi@tuebingen.mpg.de}\\
     \Name{Vincent Berenz$^{2}$} \Email{vberenz@tuebingen.mpg.de}\\
  \Name{Vaibhav Agrawal$^{2}$} \Email{vaibhav.agrawal@tuebingen.mpg.de}\\
   \Name{Bernhard Sch\"olkopf$^{2}$} \Email{bs@tuebingen.mpg.de} \begin{singlespace}  
  \addr \hskip-1.5em $^{*}$Equal contribution \\ $^{1}$ KTH Stockholm \\ $^{2}$ MPI for Intelligent Systems \\ $^{3}$ MILA \end{singlespace} 
 \AND
    \Name{Niklas Funk$^{\dag}$$^{3}$}, \Name{Julen Urain De Jesus$^{\dag}$$^{3}$},  \Name{Jan Peters$^{\dag}$$^{3}$},  \Name{Joe Watson$^{\dag}$$^{3}$},  \Name{Claire Chen$^{\dag}$$^{4}$},  \Name{Krishnan Srinivasan$^{\dag}$$^{4}$},  \Name{Junwu Zhang$^{\dag}$$^{4}$},  \Name{Jeffrey Zhang$^{\dag}$$^{4}$},  \Name{Matthew R. Walter$^{\dag}$$^{5}$},  \Name{Rishabh Madan$^{\dag}$$^{9}$},  \Name{Charles Schaff$^{\dag}$$^{5}$},   \Name{Takuma Yoneda$^{\dag}$$^{5}$},  \Name{Denis Yarats$^{\dag}$$^{6}$},  \Name{Arthur Allshire$^{\dag}$$^{7}$},  \Name{Ethan K. Gordon $^{\dag}$$^{8}$},   \Name{Tapomayukh Bhattacharjee$^{\dag}$$^{9}$},  \Name{Siddhartha S. Srinivasa$^{\dag}$$^{8}$}, \Name{Animesh Garg$^{\dag}$$^{7}$}, \Name{Takahiro Maeda$^{\dag}$$^{10}$}, \Name{Harshit Sikchi$^{\dag}$$^{11}$}, \Name{Jilong Wang$^{\dag}$$^{12}$}, \Name{Qingfeng Yao$^{\dag}$$^{12}$}, \Name{Shuyu Yang$^{\dag}$$^{12}$}, \Name{Robert McCarthy$^{\dag}$$^{13}$}, \Name{Francisco Roldan Sanchez$^{\dag}$$^{14}$}, \Name{Qiang Wang$^{\dag}$$^{13}$}, \Name{David Cordova Bulens$^{\dag}$$^{13}$}, \Name{Kevin McGuinness$^{\dag}$$^{14}$}, \Name{Noel O'Connor$^{\dag}$$^{14}$}, \Name{Stephen J. Redmond$^{\dag}$$^{13}$}\\ 
  \addr \newline $^{\dag}$ Challenge participant, $^{3}$TU Darmstadt, $^{4}$Stanford University, $^{5}$TTI Chicago, $^{6}$New York University, $^{7}$University of Toronto, $^{8}$University of Washington, $^{9}$Cornell University, $^{10}$ TTI, $^{11}$ The University of Texas, $^{12}$ Westlake University, $^{13}$ University College Dublin, $^{14}$ Dublin City University \\
 }

\begin{document}

\maketitle
\newpage
\begin{abstract}
Dexterous manipulation remains an open problem
in robotics.
To coordinate efforts of the research community towards tackling this problem, we propose a shared benchmark. 
We designed and built robotic platforms that are hosted at the MPI-IS$^{1}$ and can be accessed remotely.
Each platform consists of three robotic fingers that are capable of dexterous object manipulation.
Users are able to control the platforms remotely by submitting code that is executed automatically, akin to a computational cluster. 
Using this setup, i) we host robotics competitions, where teams from anywhere in the world access our platforms to tackle challenging tasks ii) we publish the datasets collected during these competitions (consisting of hundreds of robot hours), and iii) we give researchers access to these platforms for their own projects.
\end{abstract}

\section{Introduction}

Dexterous manipulation is humans' interface to the physical world. Our ability to
manipulate objects around us in a creative and precise manner is
one of the most apparent distinctions between human and animal intelligence. The
impact robots with a similar level of dexterity would have on our society cannot
be overstated. They would likely replace humans in most tasks that are primarily
physical, such as working at production lines, packaging, constructing houses, agriculture, cooking, and cleaning.
Yet, robotic manipulation is still far from the level of dexterity attained by 
humans, as witnessed by the fact that these are still mostly 
carried out by humans. This problem has been remarkably resistant to the 
rapid progress of machine learning over the past years.
A factor that has been crucial for progress in machine learning, but 
nonexistent in real-world robotic manipulation, is a shared benchmark. 
Benchmarks allow for different labs to coordinate efforts, reproduce results and measure progress. Most notably, in the area 
of image processing, such benchmarks were crucial for the rapid progress of deep learning. More recently, simulation benchmarks have been proposed in reinforcement learning (RL) \cite{brockman2016openai, tassa2018deepmind}.
However, methods that are successful in simulators transfer only to a limited
degree to real robots.
Therefore, the robotics community has recently proposed a number of open-source platforms for robotic manipulation \cite{yang2019replab, ahn2019robel, Wuthrich2020-by}. These 
robots can be built by any lab to reproduce results of other labs. 
While
this is a large step towards a shared benchmark, it requires 
effort by the researchers to set up and maintain the system, and it is 
nontrivial to ensure a fully standardized setup.

Therefore, we provide \textbf{remote access to dexterous
manipulation platforms} hosted at MPI-IS, see \Cref{fig:robots} in the appendix and  \Cref{fig:challenge_tasks} (\emph{interested researchers can contact us to request access, see our website}\footnote{\label{trifinger_site}\url{https://people.tuebingen.mpg.de/felixwidmaier/trifinger}}).
This allows for an objective evaluation of robot-learning algorithms on real-world platforms with
minimal effort for the researchers. In addition, we publish a \textbf{large
dataset of these platforms interacting with objects}, which we collected during
a competition we hosted in 2020 and 2021 as part of the Neural Information Processing Systems (NeurIPS) Conference\footnote{\url{https://real-robot-challenge.com/2020}}. During this competition, teams from
across the world developed algorithms for challenging object manipulations
tasks, which yielded a very diverse dataset containing meaningful interactions. 

%
%
To facilitate research into sim-to-real transfer, we also provide a \textbf{simulation of the robotic setup} (see website\footnote{same as footnote \ref{trifinger_site}}).
All the code, for both simulation and control of the real robots, is open-source.



In the rest of the paper, we describe the robotic hardware and the software interface which allows
easy robot access, similarly to a computational cluster. 
We also describe the robot competitions we hosted and the data 
we collected in the process.

\section{Related Work}

In the past years, a large part of the RL community has focused on simulation
benchmarks, such as the deepmind control suite \cite{tassa2018deepmind} or
OpenAI gym \cite{brockman2016openai} and extensions thereof
\cite{zamora2016extending}. These benchmarks internally use physics simulators,
typically Mujoco \cite{todorov2012mujoco} or PyBullet
\cite{coumans2016pybullet}.

These commonly accepted benchmarks allowed researchers from different labs to
compare their methods, reproduce results, and hence build on each other's work.
Very impressive results have been obtained through this coordinated effort
\cite{Haarnoja2018-ox, Fujimoto2018-ur, Popov2017-hh, Mnih2016-mn,
Heess2017-vd, duan2016benchmarking, henderson2018deep}.

In contrast, no such coordinated effort has been possible on real robotic systems,
since there is no shared benchmark.
This lack of standardized real-world benchmarks has been recognized by the
robotics and RL community \cite{behnke2006robot,
Bonsignorio2015-lg, Calli2015-vu, Calli2015-zj, Amigoni2015-gh, Murali2019-rg}.
Recently, there have been renewed efforts to alleviate this problem:

\paragraph{Affordable Open-Source Platforms:}
The robotics community recently proposed affordable open-source robotic platforms that can be built by users. For instance, \cite{yang2019replab} propose Replab, a simple, low-cost manipulation
platform that is suitable for benchmarking RL algorithms. Similarly, \cite{ahn2019robel} propose a simple robotic hand and quadruped that can be built from commercially available modules. CMU designed LoCoBot, a
low-cost open-source platform for mobile manipulation. 
\cite{Grimminger2020-tl} propose an
open-source quadruped consisting of off-the-shelf parts and 3D-printed shells. Based on this design, \cite{Wuthrich2020-by} developed an open-source manipulation platform consisting of three fingers capable of complex dexterous manipulation (here, we use an industrial-grade adaptation of this design).

Such platforms are beneficial for collaboration and reproducibility across labs. However, setting up and maintaining such platforms often requires hardware experts and is time-intensive. Furthermore, there are necessarily small variations across labs that may harm reproducibility. To overcome these limitations, the robotics community has proposed a number of remote robotics benchmarks.

\paragraph{Remote Benchmarks:}
For mobile robotics, \cite{pickem2017robotarium} propose the Robotarium, a remotely accessible swarm robotics research platform.
Similarly, Duckietown \cite{Paull2017-so} hosts the AI Driving Olympics \cite{AI-DO_AI_Driving_Olympics_Duckietown_undated-mw} twice per year. However, a remote benchmark for robotic manipulation accessible to researchers around the world is still missing. Therefore we propose such a system herein.

\section{Robotic Platforms}\label{sec:robotic_platforms}
We host 8 robotic platforms at MPI-IS (see  \Cref{fig:challenge_tasks,fig:robots}), remote users can submit code which
is then assigned to a platform and executed automatically, akin to a
computational cluster. Users have access to the data 
collected during execution of their code. Submission and data retrieval can be automated to allow for RL methods that alternate between policy evaluation and policy improvement.

The platforms we use here are based on an open-source design that was published 
recently \cite{Wuthrich2020-by} (see also website\footnote{\url{https://sites.google.com/view/trifinger}}). The benefits of this design are 
\begin{itemize}
 \setlength\itemsep{0em}
	\item\textbf{Dexterity:} The robot design consists of three fingers and has
	the mechanical and sensorial capabilities necessary for complex object
	manipulation beyond grasping.
	\item \textbf{Safe Unsupervised Operation:} The combination of robust hardware
	and safety checks in the software allows users to run even unpredictable
	algorithms without supervision. This enables, for instance, training of deep
	neural networks directly on the real robot.
	\item \textbf{Ease of Use:} The C++ and Python interfaces
	are simple and well-suited for RL as well as optimal control
	at rates up to \SI{1}{kHz}. For convenience, we also provide a simulation
	(PyBullet) environment of the robot.
\end{itemize}

Here, we use an industrial-grade adaptation of this hardware (see \Cref{fig:challenge_tasks,fig:robots}) to
guarantee an even longer lifetime and higher reproducibility. In addition, we developed a submission system to allow
researchers from anywhere in the world to submit code with ease.

\subsection{Observations and Actions}
\label{sec:observations_and_actions}

\textbf{Actions:} This platform consists of 3 fingers, each with 3 joints, 
yielding a total of 9 degrees of freedom (and 9 corresponding motors). 
There are two ways of controlling the robot:
One can send 9-dimensional \textbf{torque-actions} which are directly executed by the 
motors. Alternatively, we provide the option of using 
\textbf{position-actions} (9-dimensional as well), which are then 
translated to torques by an internal controller.
The native control rate of the system is 1kHz, but one can control 
at a lower rate, if so desired.

\textbf{Observations:} An observation consists of proprioceptive measurements,
images and the object pose inferred using an object tracker. The
proprioceptive measurements are provided at a rate of 1\,kHz and contain the \textbf{joint angles, joint velocities, joint
torques} (each 9 dimensional) and \textbf{finger-tip forces} (3 dimensional).
There are three cameras placed around the robot.
The \textbf{camera images} are provided at 10\,Hz and have a resolution of 
270x270 pixels. In addition, our system also contains an object
tracker, which provides the \textbf{pose of the object} being manipulated along 
with the images.



\subsection{Submitting Code}
\label{sec:submission_system}

We use HTCondor\footnote{\url{https://research.cs.wisc.edu/htcondor}} to provide
a cluster-like system where users can submit jobs which are then automatically
executed on a randomly-selected robot. During execution, a backend process
automatically starts the robot, monitors execution and records all actions and
observations.  The users use the simple front-end interface to send actions to
and retrieve observations from the robot (see \cite{Wuthrich2020-by} for more details on the design of the interface, see the website\footnote{\url{https://people.tuebingen.mpg.de/felixwidmaier/trifinger}} for links to the code repositories).

Below is a minimal example of actual user code in Python. It creates a
front-end interface to the robot and uses it to send torque commands that are
computed by a user-specified control policy based on the observations:

\begin{lstlisting}[language=Python]
# Initialise front end to interact with the robot.
robot = robot_fingers.TriFingerPlatformFrontend()

# Create a zero-torque action to start with
action = robot_interfaces.trifinger.Action()

while True:
    # Append action to the "action queue". Returns the time step at which the given action will be executed.
    t = robot.append_desired_action(action)

    # Get observations of time step t. Will wait if t is in the future.
    robot_obs = robot.get_robot_observation(t)
    camera_obs = robot.get_camera_observation(t)

    # Compute next action using some control policy. The different observation values are listed separately for illustration.
    torque = control_policy(
        robot_obs.position,
        robot_obs.velocity,
        robot_obs.torque,
        robot_obs.tip_force,
        camera_obs.cameras[0].image,
        camera_obs.cameras[1].image,
        camera_obs.cameras[2].image,
        camera_obs.object_pose,
    )
    action = robot_interfaces.trifinger.Action(torque=torque)
\end{lstlisting}
%
%
%
At the end of each job, the recorded data is stored and provided to the user, who can then analyse it and use it, for example, to train a better policy. Users can automate submissions and data retrieval to run RL algorithms directly on the robots.
See our website\footnote{\url{https://people.tuebingen.mpg.de/felixwidmaier/trifinger}} to learn how to get started and submit code to our robots.

\subsection{Simulation}
To facilitate testing code and sim-to-real transfer, we provide a simulation of the platform with same software interface as the real robot\footnote{\url{https://open-dynamic-robot-initiative.github.io/trifinger_simulation}}.
In addition, we developed a more advanced version of the simulator where parameters of the robot and the environment (such as masses, colors of objects, size of objects etc.) can be changed easily (see paper \cite{causalworld} and website\footnote{\url{https://sites.google.com/view/causal-world/home}}). This allows for learning the causal structure of the control problem and facilitates transfer learning, in particular transferring a policy to the real world.

\section{The Real Robot Challenge}

Using these TriFinger robots, we organized two competitions, called "Real Robot Challenge" (RRC), in 2020 and in 2021 as part of NeurIPS.
In this section we describe the general structure of the challenges and how user submissions were evaluated.  The specific tasks and results of each challenge will be presented in \Cref{sec:rrc2020,sec:rrc2021}.

\subsection{Challenge Phases}
\label{sec:challenge_phases}

Each challenge was split into three phases.
The first phase consisted of a task in our simulator and served as a qualification round before granting access to the real robots.
Teams who achieved promising results in the first phase could then move on to phases 2 and 3 on the real robots.  They were given remote access through the submission system described in \Cref{sec:submission_system}.
Phase 2 consisted of solving the task from phase 1 on the real robots, so participants could build on what they already achieved in simulation.  In the last phase the object (and in RRC 2021 also the task itself) was changed.

\subsection{Evaluation}
\label{sec:challenge_evaluation}

Each task had to be solved within a fixed time frame (usually an episode of two minutes, corresponding to 120000 steps at 1\,kHz).  For the evaluation, a task-specific reward $r_t$ was computed for every time step $t$. We defined the performance of an episode as the cumulative reward $R = \sum_t r_t$. 
At the end of each real-world phase the code of each team was executed for multiple goals on different robots (using the same set of goals for all users). The average cumulative reward of these runs was then used to rank the submissions.

\begin{figure}[ht!]
    \centering
    \subfigure[Simulation]{
        \includegraphics[width=0.4\textwidth]{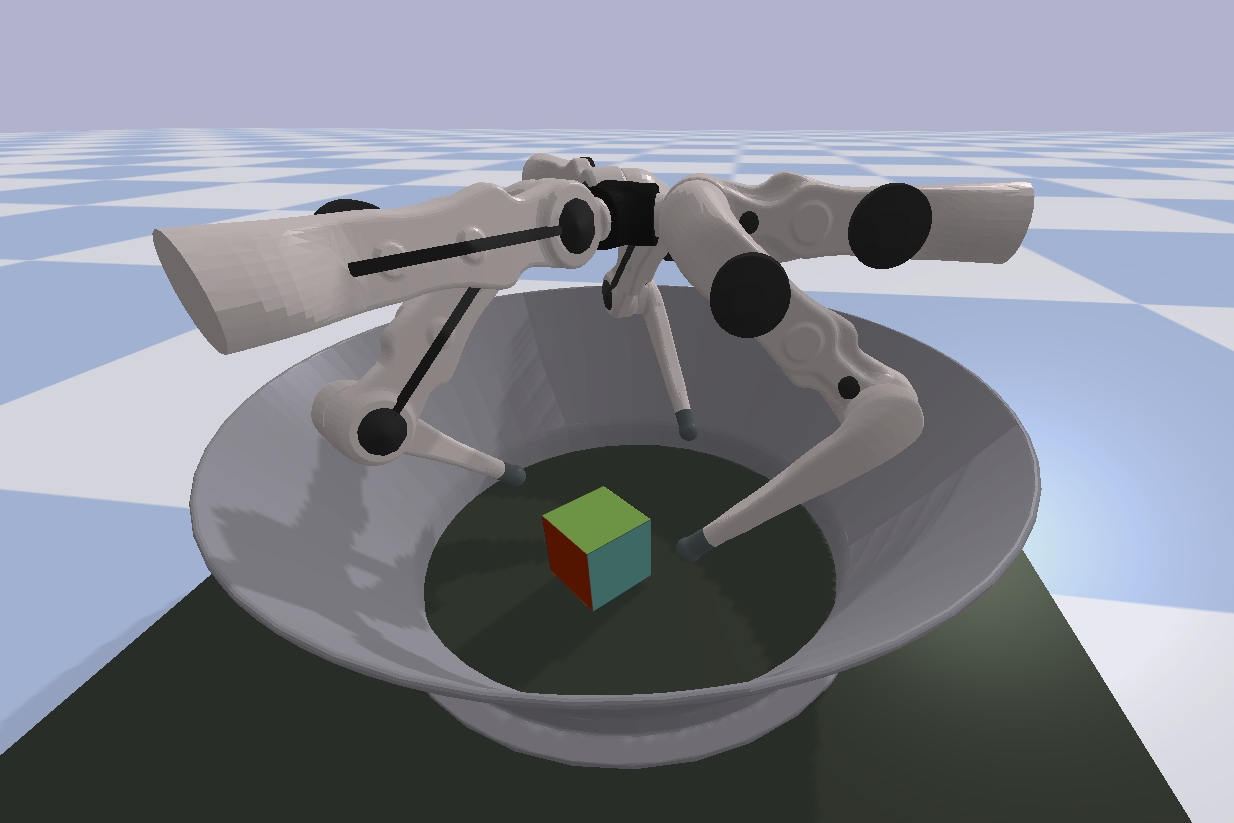}
    }
    \subfigure[Cube]{
        \includegraphics[width=0.4\textwidth]{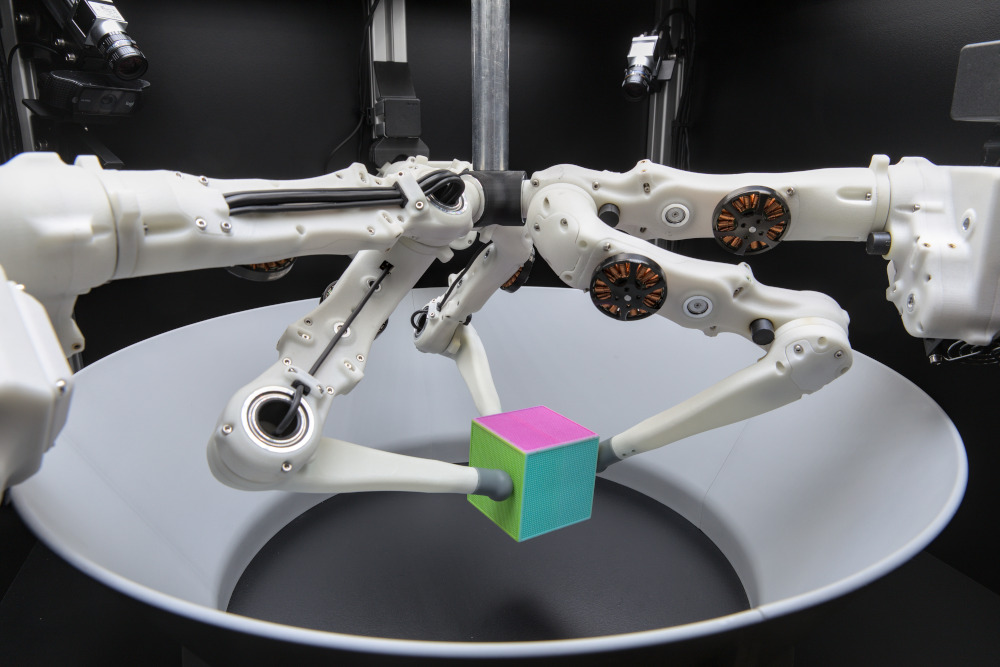}
    }
    \subfigure[Elongate Cuboid (only 2020)]{
        \includegraphics[width=0.4\textwidth]{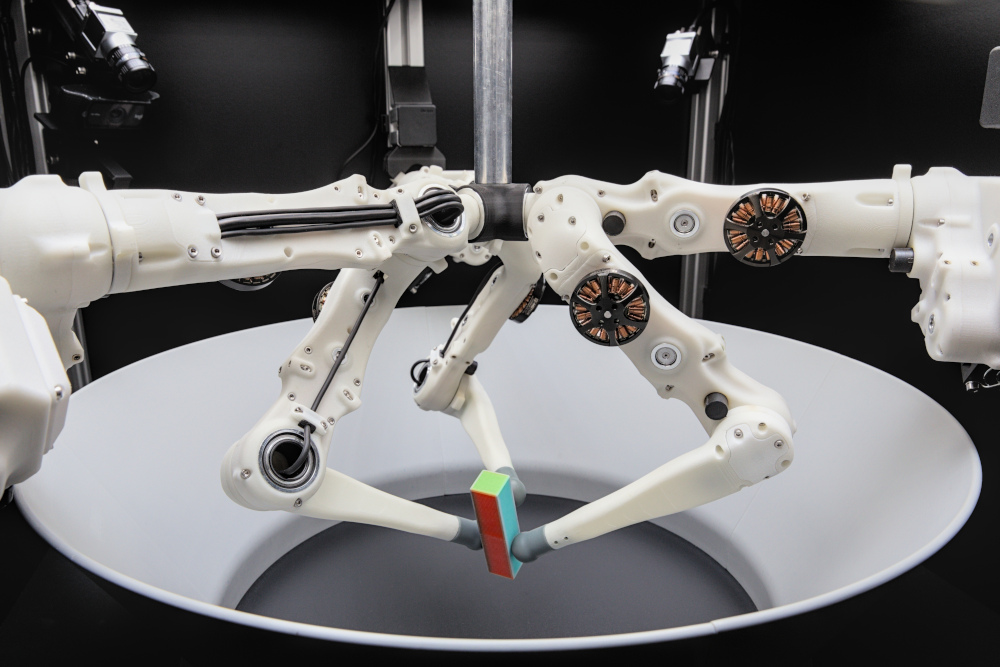}
    }
    \subfigure[Dice (only 2021)]{
        \includegraphics[width=0.4\textwidth]{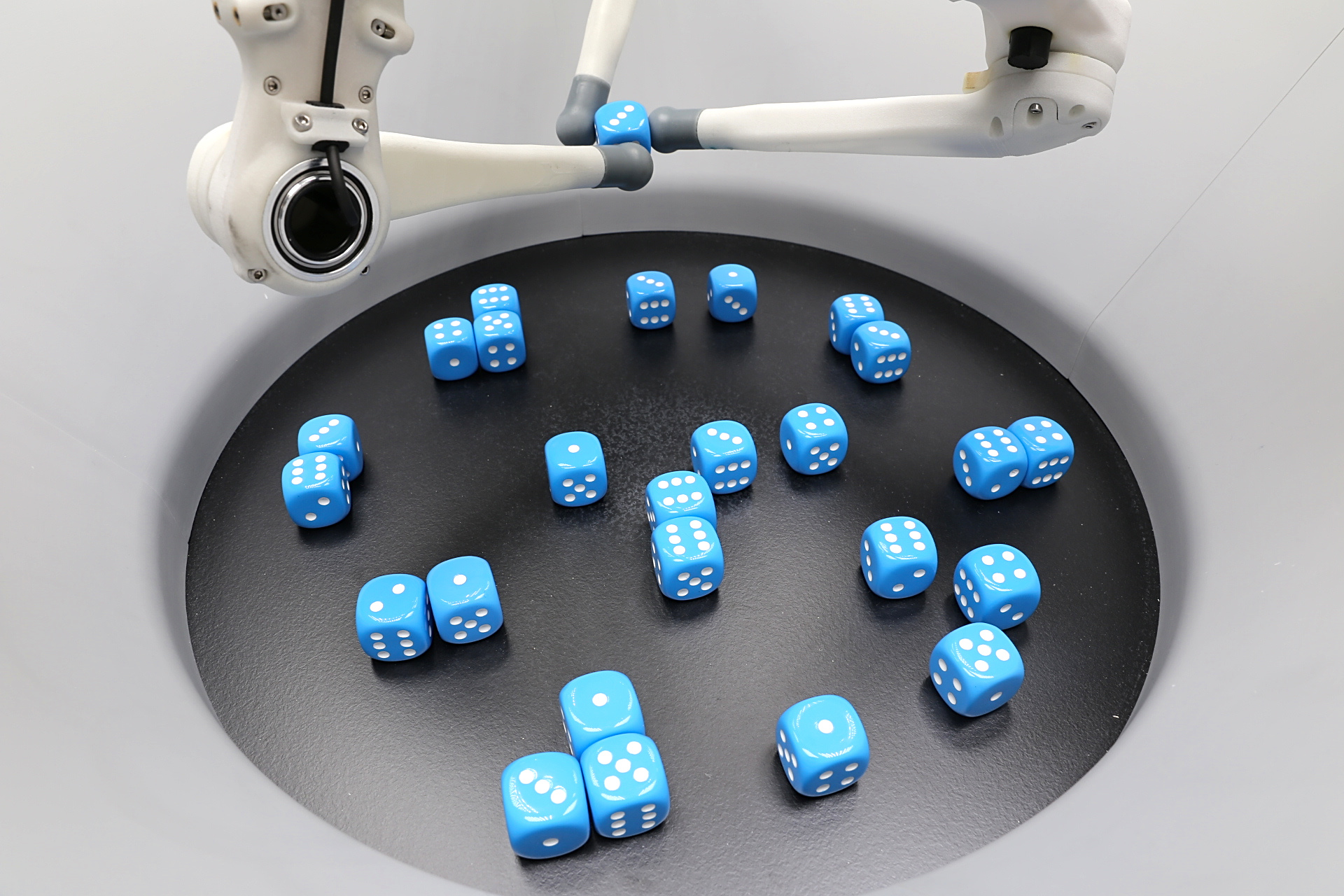}
    }
    \caption{Illustration of the different challenge tasks.  In RRC 2020 the
    cube and the cuboid were used and in RRC 2021 the cube and the dice.}
    \label{fig:challenge_tasks}
\end{figure}

\section{Tasks and Results of RRC 2020}
\label{sec:rrc2020}

The first challenge lasted from August to December 2020.
During the challenge, the data of all runs made by the participants was recorded, resulting in a large dataset that is rich in contact interactions.  This dataset is publicly available and also describe in the following.

\subsection{Tasks}
\label{sec:rrc2020_tasks}

In all three phases, the task was to move an object from its initial position
at the center of the workspace to a randomly sampled goal.  In phases 1
(simulation) and 2 this object was a 65\,mm cube, in phase 3 it was replaced by
a smaller, elongate cuboid (20$\times$20$\times$80\,mm) that is more difficult to grasp and
manipulate. \Cref{fig:challenge_tasks} shows pictures of the different objects.
In each phase, there were four levels of difficulty corresponding to different
goal distributions:
\begin{itemize}
 \setlength\itemsep{0em}
    \item \textbf{Level 1:} The goal is randomly sampled on the table, so it can be attained by pushing the object. The
      orientation is not considered for the reward computation.
    \item\textbf{Level 2:} The object has to be lifted to a fixed goal position
      8\,cm above the table center. The orientation is not considered for the reward computation.
    \item\textbf{Level 3:} The goal is randomly sampled somewhere within the
      arena with an height of up to 10\,cm. The orientation is not considered for the reward computation.
    \item\textbf{Level 4:}  As level 3, but in addition to the position, a goal
      orientation is sampled uniformly.
\end{itemize}

\subsubsection{Reward Function for Levels 1-3}\label{sec:reward_level123}

For difficulty levels 1-3, we only considered position error (i.e. orientation is ignored).
We used a weighted sum of the Euclidean distance on the x/y-plane and the
absolute distance along the z-axis. Both components are scaled based on their
expected range. The sum is again rescaled so that the total error is in the
interval $[0, 1]$.
Given goal position $p_g = (x_g, y_g, z_g)$, actual position $p_a = (x_a, y_a,
z_a)$, arena diameter $d$ and maximum expected height $h$, the position error
$e_\mathrm{pos}$ is computed as
\begin{equation}\label{eq:position_error}
    e_\mathrm{pos} = \frac{1}{2} \left( \frac{\sqrt{(x_g - x_a)^2 + (y_g - y_a)^2}}{d} + \frac{|z_g - z_a|}{h} \right)
\end{equation}
We set $d = 0.39$\,m (matching the inner diameter of the arena boundary), and $h = 0.1$\,m.
The reward $r$ is then simply the negative error $r = -e_\mathrm{pos}$.

\subsubsection{Reward Function for Level 4}
\paragraph{Phases 1 and 2 (Cube)}


For level 4, we considered both position and orientation.  
The position error $e_\mathrm{pos}$ is computed as in the previous level, according to \cref{eq:position_error}.
We define the rotation error as the normalized angle of the rotation (represented as quaternion $q$) that would have to be applied to the orientation of the object to match the goal orientation:
\begin{equation}
    e_\mathrm{rot} = \frac{2 \cdot \operatorname{atan2}(\|(q_x, q_y, q_z)\|, |q_w|)}{\pi}.
\end{equation}
%
As reward we use the negative average error $r = - \frac{e_\mathrm{pos} + e_\mathrm{rot}}{2}$, which lies in the interval $[0, 1]$.

\paragraph{Phase 3 (Cuboid)}

We found that for the narrow cuboid used in phase 3, our object tracking method
was unreliable with respect to the rotation around the long axis of the cuboid.
To prevent this from affecting the reward computation, we changed the
computation of the rotation error to use only the normalized absolute angle between the
long axes of the cuboid in goal and actual pose.


\subsubsection{Evaluating Submissions}
As described in \Cref{sec:challenge_evaluation}, we used the average score over
multiple runs with different goals for evaluating the submissions.
Since there were multiple difficulty levels for the goals, an equal number of
goals of each level $i$ was used and the average cumulative Reward $R_i$ was
computed separately for each level.
The total score for the ranking was then computed as a weighted sum
$\sum_{i=1}^4 i \cdot R_i$. This gives a higher weight to higher (more
difficult) levels, encouraging teams to solve them.

\subsection{Challenge Results}

After the simulation stage, seven teams with excellent performance qualified for
the real-robot phases. These teams made thousands of submissions to the robots,
corresponding to approximately 250 hours of robot-run-time.
See \appendixref{appendix:results_and_submission_statistics} for the final evaluation results and submission statistics of phases 2 and 3.
The top teams in both phases found solutions that successfully grasp the object
and move it to the goal position.  Videos published by the winning teams are
available on YouTube\footnote{ardentstork: \url{https://youtube.com/playlist?list=PLBUWL2_ywUvE_czrinTTRqqzNu86mYuOV}\\
troubledhare: \url{https://youtube.com/playlist?list=PLEYg4qhK8iUaXVb1ij18pVwnzeowMCpRc} \\
sombertortoise: \url{https://youtu.be/I65Kwu9PGmg}}.
They also published reports describing their methods
(\cite{yoneda2021grasp,troubledhare2020real,chen2021dexterous}) and open-sourced
their code\footnote{ardentstork: \url{https://github.com/ripl-ttic/real-robot-challenge} \\
troubledhare: \url{https://github.com/madan96/rrc_example_package} \\
sombertortoise: \url{https://github.com/stanford-iprl-lab/rrc_package}}.
We collected all the data produced during the challenge and aggregated it into a dataset, which is described in \Cref{sec:dataset}.

\paragraph{The Winning Policies}

The winning teams used similar methods for solving the task:  They made use of
motion primitives that are sequenced using state machines.  For difficulty level
4, where orientation is important, they typically first perform a sequence of
motions to rotate the object to be roughly in the right orientation before
lifting it to the goal position.
For details regarding their implementations, please refer to
\cite{yoneda2021grasp,troubledhare2020real,chen2021dexterous,
allshire2021transferring}.  Further \cite{funk2021benchmarking} contains a more
detailed description of the solutions of some of the teams.




\subsection{The Dataset}
\label{sec:dataset}

The dataset contains the recorded data of all jobs that were executed during phases 2 and 3 of the challenge.
Combined with the runs from the weekly evaluation rounds, this results in 2856 episodes of phase 2 and 7422 episodes of phase 3.
The data of each episode can be downloaded individually.
It contains all robot and camera observations (including object pose information) as well as all actions that were sent by the user (see \Cref{sec:observations_and_actions}), the goal pose that was used in this episode, camera calibration parameters and metadata such as timestamp, challenge phase, etc.
Further some metrics like cumulative reward, initial distance to goal, maximum height of the object throughout the episode and some more.
We expect that users of the dataset will typically not use all episodes (which
would be a large amount of data), but select those that are interesting for
their project.  To help with this, we provide a database containing the metadata and metrics of all episodes. This allows to filter the data before downloading.
The dataset itself, the tools for filtering the episodes as well as a more
technical description of the data format can be found on the dataset website\footnote{\url{https://people.tuebingen.mpg.de/mpi-is-software/data/rrc2020}}.

\section{Tasks and Results of RRC 2021}
\label{sec:rrc2021}

After the successful RRC 2020, we organised a second challenge from May to
September 2021.  The challenge was structured in the same way as the previous
one but with new tasks.

\paragraph{Note on nomenclature ("phase" vs "stage"):} While in 2020 the phases
were called \emph{phase 1,2,3}, the naming changed to \emph{pre-stage} and
\emph{stage 1,2} in 2021.
We keep the respective wording for each challenge, to be consistent with documentation and other publications.

\subsection{Tasks}
\label{sec:rrc2021_tasks}

In the simulation-stage ("pre-stage") and in stage 1 the \emph{move cube on trajectory} task had to be solved, in stage 2 the \emph{rearrange dice} task.
\Cref{fig:challenge_tasks} shows pictures of the different objects.

\subsubsection{Move Cube on Trajectory}

The \emph{move cube on trajectory} task is an extension of the task of RRC 2020
using the 65\,mm cube (see \Cref{sec:rrc2020_tasks}).
Instead of a single goal position, a list of sub-goals is given with the
active sub-goal changing over time. So the cube has to be moved on a trajectory
from goal to goal.
The active sub-goal changes every 10 seconds, with an exception on the first
one, which is active for 30 seconds to account for the additional time needed to
pick up the cube.
Each sub-goal is sampled independently somewhere within the arena.  Only the
position is considered, so the reward is computed as in \Cref{sec:reward_level123}.

\subsubsection{Rearrange Dice}

For the \emph{rearrange dice} task the arena is filled with 25 dice (regular
six-sided dice with a width of 22\,mm), which are initially shuffled around
randomly.
They have to be arranged in a given 2d pattern, consisting of 25 goal positions
on the ground of the arena.

To evaluate a given state, a "goal mask" is created by projecting bounding cubes
at the goal positions into the camera images. 
The error $e$ is computed as the number of pixels in the segmentation mask (i.e. the pixels in which dice are visible, determined from color segmentation) that
are not in the goal mask.
Let $G_c$ the set of pixels of the goal mask and $S_c$ the pixels of the
segmentation mask of camera $c \in \{1,2,3\}$.
\begin{equation}
  e = \sum_{c \in \{1,2,3\}} |S_c \setminus G_c|
  \label{eq:rearrange_dice_cost}
\end{equation}
As for the other tasks, the reward is then simply the negated error: $r = -e$.

\subsection{Challenge Results}

After the pre-stage in simulation, six teams qualified for the real-robot
stages.  Four of the six teams submitted a solution in the end of stage 1 (see \Cref{appendix:results_and_submission_statistics} for evaluation results and submission statistics).

It appears that the task for stage 2 was too difficult, since none of the teams managed to solve it, therefore we only discuss the results of stage 1 in the following.

\paragraph{The Winning Policies of Stage 1}

Compared to the RRC 2020, the approaches were much more diverse in 2021:
\begin{itemize}
 \setlength\itemsep{0em}
  \item The winning team \textbf{thriftysnipe} \cite{mccarthy2021solving} used
    pure reinforcement learning (DDPG + HER) with minimal domain-specific
    knowledge.
  \item Team \textbf{decimalswift} \cite{decimalswift2021real} extended the
    CPC-TG approach from RRC 2020 \cite{funk2021benchmarking} (using
    position-based motion primitives for the finger tips) by interpolating
    trajectories between the goal positions for smoother motions.
  \item Team \textbf{grumpyzebra} \cite{yao2021realworld} used a mixed approach:
    They employ reinforcement learning to find good contact points on the cube
    and then use classic position control to move the finger tips.
\end{itemize}
Most teams published videos on YouTube\footnote{thriftysnipe: \url{https://www.youtube.com/playlist?list=PLLJoWXUn8XplFszi16-VZMTDBhMQFuc5o}\\
decimalswift: \url{https://youtu.be/dlOueoaRWrM}, 
grumpyzebra: \url{https://youtu.be/Jr176xsn9wg}}
and open-sourced their code\footnote{thriftysnipe: \url{https://github.com/RobertMcCarthy97/rrc_phase1}\\
grumpyzebra: \url{https://github.com/42jaylonw/RRC2021ThreeWolves}}.

\section{Takeaways from the Challenges}

The following are a few interesting insights we gained from organizing these two challenges.
\paragraph{Reinforcement Learning Versus Classical Control:} %
Interestingly, the most successful teams in the challenge of 2020 relied on traditional approaches to control rather than machine learning. In particular, hand-crafted state machines and motion primitives proved to perform well.
However, the picture was more diverse in the challenge of 2021, where in stage 1 the winning team (thriftysnipe) solved the task using reinforcement learning and the runner-up team
(decimalswift) employed an extension of a classical, control-based approach from the
previous year. It is instructive to compare how these two completely different approaches perform: The classic approach from decimalswift achieves a stable grip of the cube and drops it rarely, but moves slowly. In contrast, the RL-based solution drops the cube more often, but it is able to recover very quickly and moves very fast. It learned that it pays off to move faster at the expense of making more mistakes, a strategy that would likely not have been found by an engineer hand-designing a policy.

\paragraph{Task Design:} It can be hard to predict how engaging and difficult a task will be for participants and what solutions it will motivate. If the task is too easy, not much is gained by solving it and participants will not be engaged for long. On the other hand, if it is too hard, participants will also lose motivation quickly. Therefore, we are particularly excited about opening the platforms up to researchers for their own research projects. This may organically give rise to tasks that become benchmarks that are shared across the community.

\paragraph{Platform Design:} The platform design proved to work very well. Participants learned quickly how to use the software interface and were able to create submissions with ease. During the hundreds of hours of operation, without any intervention from our side, there were almost no software and hardware failures. We addressed the rare issues that came up to make the system even more robust. Therefore, we will continue to use this setup to organize further challenges and to provide users with robot-access for their own research.

\section{Conclusion}

We designed and built a robot cluster to facilitate reproducible research into
dexterous robotic manipulation. We believe that this cluster can greatly enhance
coordination and collaboration between researchers all across the world. We
hosted competitions on these robots to advance the state-of-the-art and to
produce a publicly-available dataset that contains contact-rich interactions between the
robots and external objects. These competitions validated our platform design, as participants were able to use the robots with ease and there were almost no hardware and software failures during the hundreds of hours of unsupervised operation.
We now open these platforms to scientists around the world for their own research project.



\bibliography{references}

\clearpage

\appendix

\section{Further Illustrations of the Robotic Platform}
\label{appendix:platform}

\begin{figure}
\centering     
\subfigure[]{\includegraphics[width=.45\textwidth]{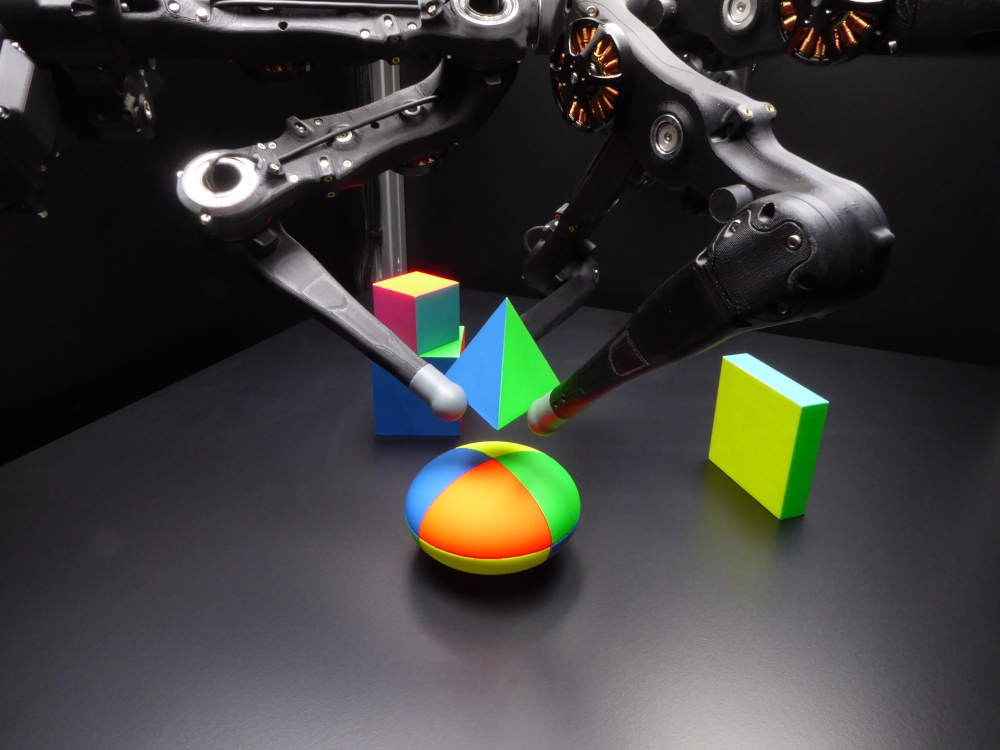}}
\subfigure[]{\includegraphics[width=.45\textwidth]{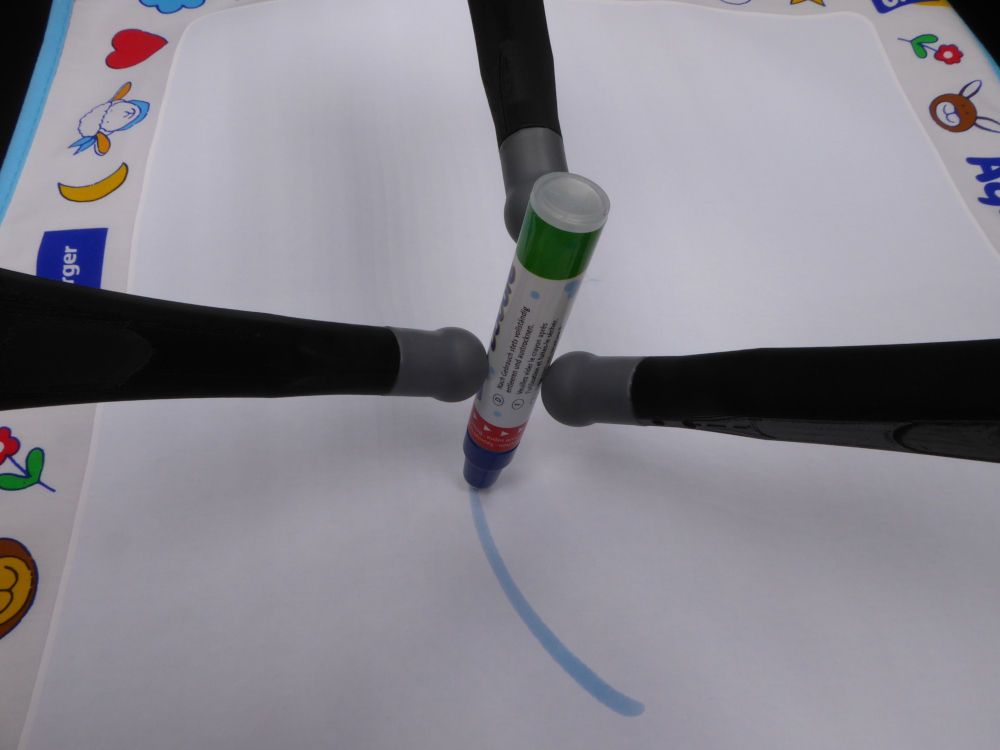}}
\subfigure[]{\includegraphics[width=.55\textwidth]{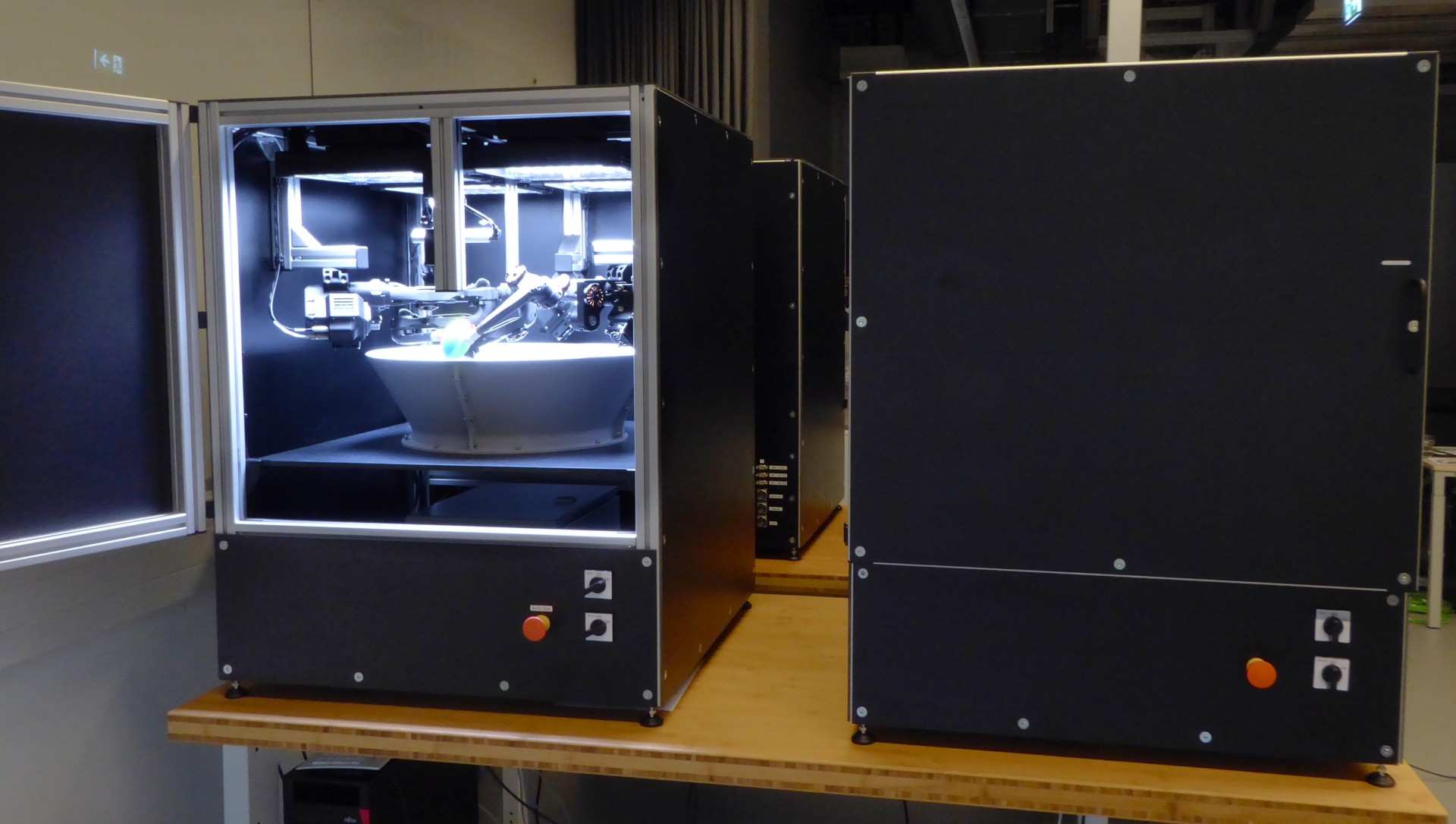}}
 \caption{The dexterous robotic platforms hosted at MPI-IS. Users can submit code which is then executed automatically.}
  \label{fig:robots}
\end{figure}

\section{Challenge Evaluation Results and Submission Statistics}
\label{appendix:results_and_submission_statistics}

\Cref{tab:ranking_phase2,tab:ranking_phase3,tab:ranking_stage1} show the results of the evaluation of submissions at the end of the real-robot phases of both challenges.
Scores in the tables are computed as described in \Cref{sec:challenge_evaluation} (see \Cref{sec:rrc2020_tasks} on how the "total score" of RRC 2020 is computed), team names correspond to the anonymous, randomly assigned names used during the challenge.

\Cref{fig:submission_statistics2020,fig:submission_statistics2021} show submission statistics of the two challenges.

\begin{table*}
\caption{Final Evaluation of RRC 2020/Phase 2}
\label{tab:ranking_phase2}
\centering
\begin{tabular}{llrrrrr}
\toprule
\# & Team & Level 1 & Level 2 & Level 3 & Level 4 & Total Score\\
\midrule
1. & ardentstork & -5472 & -2898 & -9080 & -21428 & -124221\\
2. & troubledhare & -3927 & -4144 & -4226 & -48572 & -219182\\
3. & sombertortoise & -8544 & -15199 & -14075 & -44989 & -261123\\
4. & sincerefish & -6278 & -13738 & -17927 & -49491 & -285500\\
5. & hushedtomatoe & -17976 & -41389 & -41832 & -60815 & -469509\\
6. & giddyicecream & -22379 & -46650 & -41655 & -61845 & -488023\\
\bottomrule
\end{tabular}
\end{table*}

\begin{table*}
\caption{Final Evaluation of RRC 2020/Phase 3}
\label{tab:ranking_phase3}
\centering
\begin{tabular}{llrrrrr}
\toprule
\# & Team & Level 1 & Level 2 & Level 3 & Level 4 & Total Score\\
\midrule
1. & ardentstork & -9239 & -4040 & -6525 & -25625 & -139394\\
2. & sombertortoise & -5461 & -8522 & -10323 & -36135 & -198016\\
3. & sincerefish & -7428 & -25291 & -26768 & -52311 & -347560\\
4. & innocenttortoise & -16872 & -31977 & -33357 & -55611 & -403344\\
5. & hushedtomatoe & -18304 & -31917 & -36835 & -60219 & -433521\\
6. & troubledhare & -18742 & -42831 & -36272 & -56503 & -439233\\
7. & giddyicecream & -33329 & -57372 & -53694 & -59734 & -548090\\
\bottomrule
\end{tabular}
\end{table*}

\begin{table*}
\caption{Final Evaluation of RRC 2021/Stage 1}
\label{tab:ranking_stage1}
\centering
\begin{tabular}{llrrrrr}
\toprule
\# & Team & Median Score\\
\midrule
1. & thriftysnipe & -11586\\
2. & decimalswift & -14714\\
3. & grumpyzebra & -29333\\
4. & dopeytacos & -35920\\
\bottomrule
\end{tabular}
\end{table*}

\definecolor{sombertortoise}{RGB}{181, 137, 0}
\definecolor{hushedtomatoe}{RGB}{203, 75, 22}
\definecolor{troubledhare}{RGB}{38, 139, 210}
\definecolor{sincerefish}{RGB}{42, 161, 152}
\definecolor{innocenttortoise}{RGB}{211, 54, 130}
\definecolor{giddyicecream}{RGB}{220, 50, 47}
\definecolor{ardentstork}{RGB}{133, 153, 0}

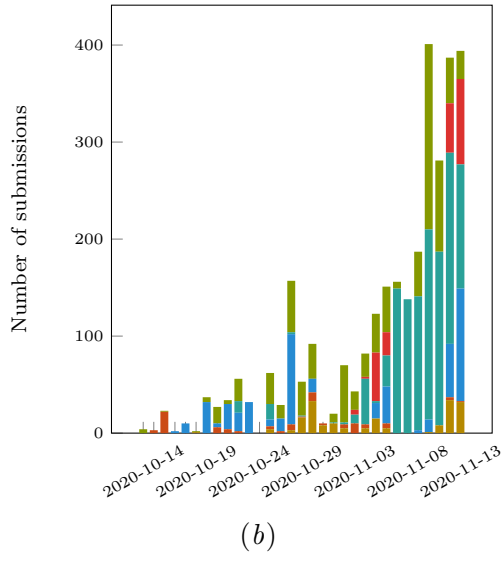
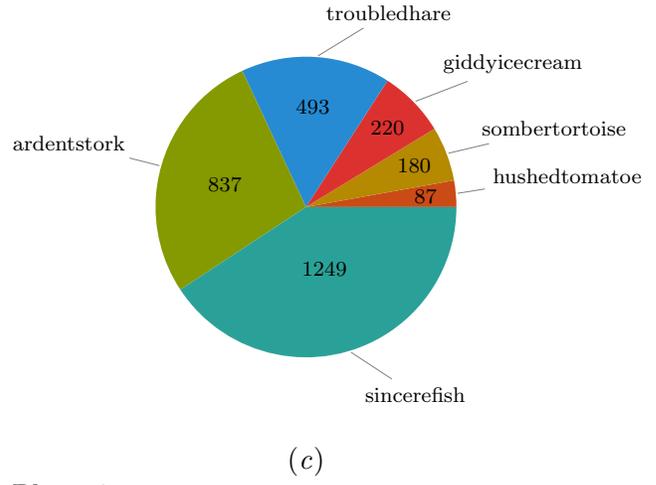
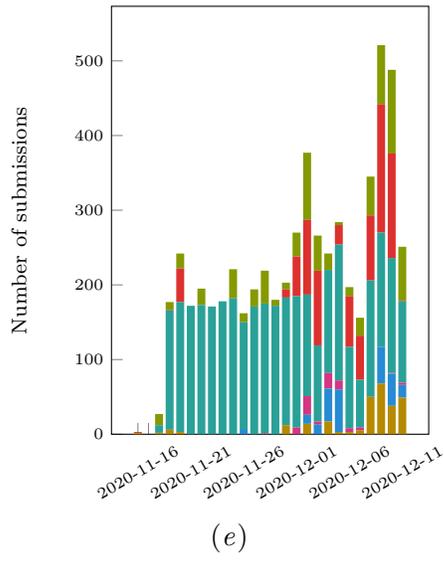
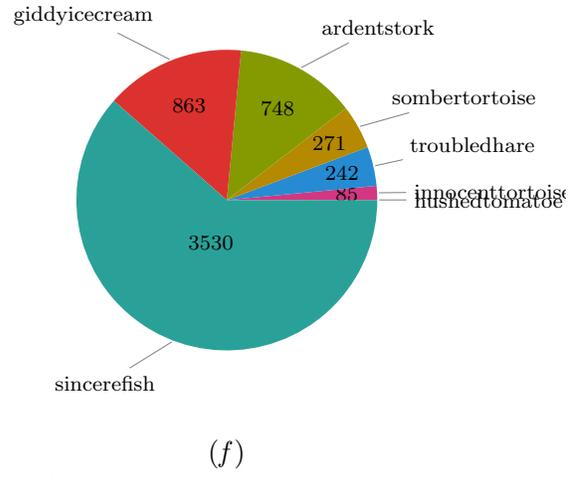
\begin{figure*}
    \centering
    \subfigure[Phase 2]{
        \centering
        \subfigure[]{
        \centering
            \begin{tikzpicture}
              \scriptsize
              \begin{axis}[
                  ybar stacked,
                  bar width=3pt,
                  x=4pt,
                  ymin=0,
                  xtick=data,
                  table/col sep=comma,
                  legend style={cells={anchor=west}, legend pos=north west},
                  reverse legend=true,
                  xticklabels from table={rrc2020_phase2_hist.txt}{Date},
                  xticklabel style={font=\tiny,rotate=30},
                  yticklabel style={font=\tiny},
                  ylabel=Number of submissions,
                  xtick pos=left,
                  ytick pos=left,
                ]
                \addplot [draw=none, fill=sombertortoise] table [y=sombertortoise, meta=Date, x expr=\coordindex] {rrc2020_phase2_hist.txt};
                \addlegendentry{sombertortoise}
                \addplot [draw=none, fill=hushedtomatoe] table [y=hushedtomatoe, meta=Date, x expr=\coordindex] {rrc2020_phase2_hist.txt};
                \addlegendentry{hushedtomatoe}
                \addplot [draw=none, fill=troubledhare] table [y=troubledhare, meta=Date, x expr=\coordindex] {rrc2020_phase2_hist.txt};
                \addlegendentry{troubledhare}
                \addplot [draw=none, fill=sincerefish] table [y=sincerefish, meta=Date, x expr=\coordindex] {rrc2020_phase2_hist.txt};
                \addlegendentry{sincerefish}
                \addplot [draw=none, fill=giddyicecream] table [y=giddyicecream, meta=Date, x expr=\coordindex] {rrc2020_phase2_hist.txt};
                \addlegendentry{giddyicecream}
                \addplot [draw=none, fill=ardentstork] table [y=ardentstork, meta=Date, x expr=\coordindex] {rrc2020_phase2_hist.txt};
                \addlegendentry{ardentstork}
                \legend{} 
              \end{axis}
            \end{tikzpicture}
        }
        \subfigure[]{
        \centering
            \begin{tikzpicture}\clip (-4.5,-3) rectangle (4.5,3);
              \scriptsize
              \tikzset{
                lines/.style={draw=none},
              }
              \pie[sum=auto, text=pin, radius=2,
                color={hushedtomatoe, sombertortoise, giddyicecream, troubledhare, ardentstork, sincerefish},
                style={lines},
              ]{87/hushedtomatoe, 180/sombertortoise, 220/giddyicecream, 493/troubledhare, 837/ardentstork, 1249/sincerefish}
            \end{tikzpicture}
        }
    }
    \subfigure[Phase 3]{
        \centering
        \subfigure[]{
        \centering
            \begin{tikzpicture}
              \scriptsize
              \begin{axis}[
                  ybar stacked,
                  bar width=3pt,
                  x=4pt,
                  ymin=0,
                  xtick=data,
                  table/col sep=comma,
                  legend style={cells={anchor=west}, legend pos=north west},
                  reverse legend=true,
                  xticklabels from table={rrc2020_phase3_hist.txt}{Date},
                  xticklabel style={font=\tiny,rotate=30},
                  yticklabel style={font=\tiny},
                  ylabel=Number of submissions,
                  xtick pos=left,
                  ytick pos=left,
                ]
                \addplot [draw=none, fill=sombertortoise] table [y=sombertortoise, meta=Date, x expr=\coordindex] {rrc2020_phase3_hist.txt};
                \addlegendentry{sombertortoise}
                \addplot [draw=none, fill=hushedtomatoe] table [y=hushedtomatoe, meta=Date, x expr=\coordindex] {rrc2020_phase3_hist.txt};
                \addlegendentry{hushedtomatoe}
                \addplot [draw=none, fill=troubledhare] table [y=troubledhare, meta=Date, x expr=\coordindex] {rrc2020_phase3_hist.txt};
                \addlegendentry{troubledhare}
                \addplot [draw=none, fill=innocenttortoise] table [y=innocenttortoise, meta=Date, x expr=\coordindex] {rrc2020_phase3_hist.txt};
                \addlegendentry{innocenttortoise}
                \addplot [draw=none, fill=sincerefish] table [y=sincerefish, meta=Date, x expr=\coordindex] {rrc2020_phase3_hist.txt};
                \addlegendentry{sincerefish}
                \addplot [draw=none, fill=giddyicecream] table [y=giddyicecream, meta=Date, x expr=\coordindex] {rrc2020_phase3_hist.txt};
                \addlegendentry{giddyicecream}
                \addplot [draw=none, fill=ardentstork] table [y=ardentstork, meta=Date, x expr=\coordindex] {rrc2020_phase3_hist.txt};
                \addlegendentry{ardentstork}
                \legend{} 
              \end{axis}
            \end{tikzpicture}
        }
        \subfigure[]{
        \centering
            \begin{tikzpicture}\clip (-4.5,-3) rectangle (4.5,3);
              \scriptsize
              \tikzset{
                lines/.style={draw=none},
              }
              \pie[sum=auto, text=pin, radius=2,
                color={hushedtomatoe, innocenttortoise, troubledhare, sombertortoise, ardentstork, giddyicecream, sincerefish},
                style={lines},
              ]{2/hushedtomatoe, 85/innocenttortoise, 242/troubledhare, 271/sombertortoise, 748/ardentstork, 863/giddyicecream, 3530/sincerefish}
            \end{tikzpicture}
        }
    }
    \caption{Number of submissions during RRC 2020 over time (left, one bar corresponds to one day) and per team (right).}
    \label{fig:submission_statistics2020}
\end{figure*}

\definecolor{worldlythrush}{RGB}{181, 137, 0}
\definecolor{thriftysnipe}{RGB}{203, 75, 22}
\definecolor{decimalswift}{RGB}{38, 139, 210}
\definecolor{grumpyzebra}{RGB}{42, 161, 152}
\definecolor{solemnlollies}{RGB}{211, 54, 130}
\definecolor{dopeytacos}{RGB}{133, 153, 0}

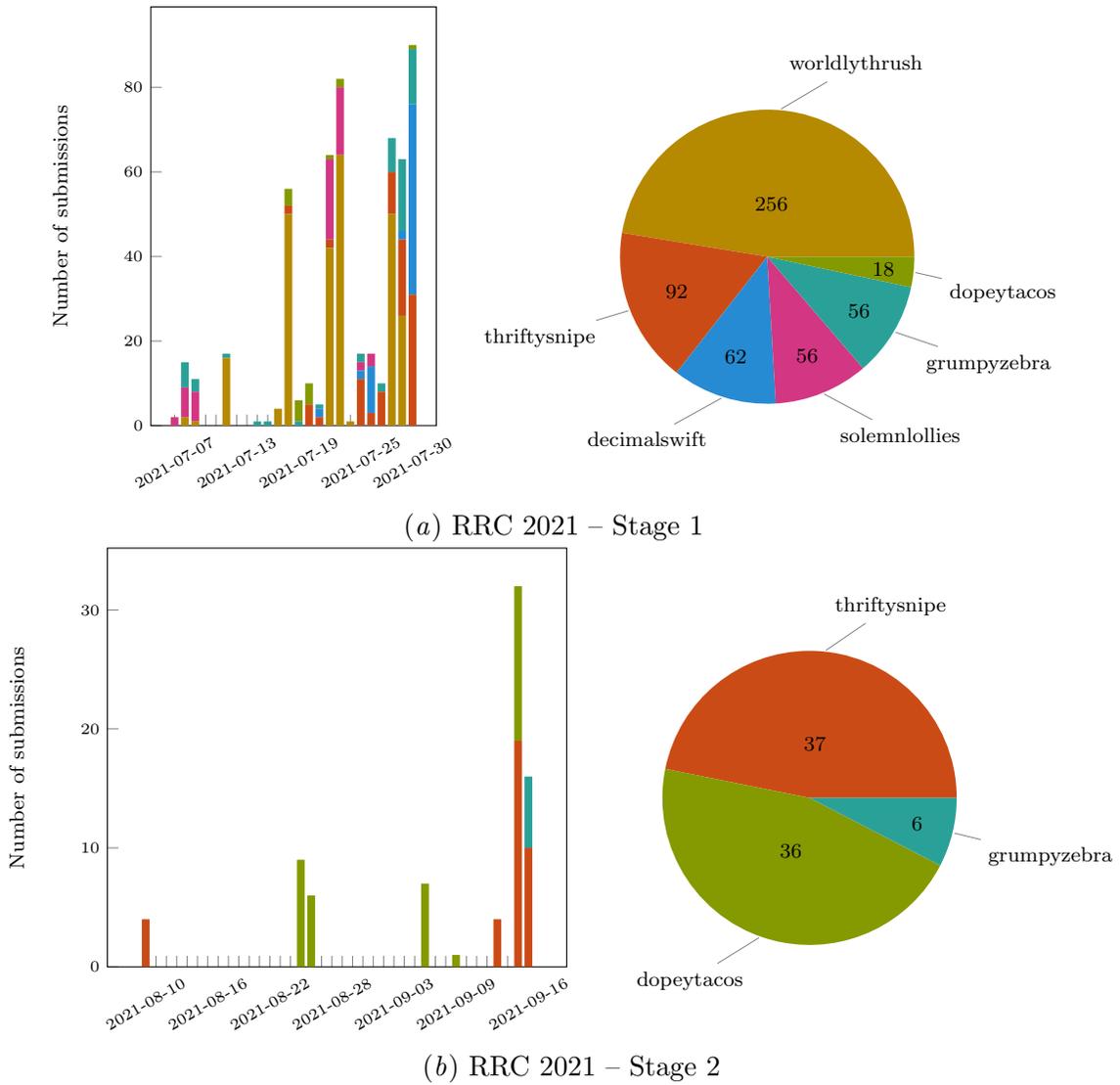
\begin{figure*}
    \centering
    \subfigure[RRC 2021 -- Stage 1]{
        \centering
        \begin{tikzpicture}
          \scriptsize
          \begin{axis}[
              ybar stacked,
              bar width=3pt,
              x=4pt,
              ymin=0,
              xtick=data,
              table/col sep=comma,
              legend style={cells={anchor=west}, legend pos=north west},
              reverse legend=true,
              xticklabels from table={rrc2021_stage1_hist.txt}{Date},
              xticklabel style={font=\tiny,rotate=30},
              yticklabel style={font=\tiny},
              ylabel=Number of submissions,
              xtick pos=left,
              ytick pos=left,
            ]
            \addplot [draw=none, fill=worldlythrush] table
              [y=worldlythrush, meta=Date, x expr=\coordindex]
              {rrc2021_stage1_hist.txt};
            \addlegendentry{worldlythrush}
            \addplot [draw=none, fill=thriftysnipe] table [y=thriftysnipe,
              meta=Date, x expr=\coordindex] {rrc2021_stage1_hist.txt};
            \addlegendentry{thriftysnipe}
            \addplot [draw=none, fill=decimalswift] table [y=decimalswift,
              meta=Date, x expr=\coordindex] {rrc2021_stage1_hist.txt};
            \addlegendentry{decimalswift}
            \addplot [draw=none, fill=solemnlollies] table
              [y=solemnlollies, meta=Date, x expr=\coordindex]
              {rrc2021_stage1_hist.txt};
            \addlegendentry{solemnlollies}
            \addplot [draw=none, fill=grumpyzebra] table [y=grumpyzebra,
              meta=Date, x expr=\coordindex] {rrc2021_stage1_hist.txt};
            \addlegendentry{grumpyzebra}
            \addplot [draw=none, fill=dopeytacos] table [y=dopeytacos,
              meta=Date, x expr=\coordindex] {rrc2021_stage1_hist.txt};
            \addlegendentry{dopeytacos}
            \legend{} 
          \end{axis}
        \end{tikzpicture}
        \begin{tikzpicture}\clip (-4,-3.3) rectangle (4,3);
          \scriptsize
          \tikzset{
            lines/.style={draw=none},
          }
          \pie[sum=auto, text=pin, radius=2,
            color={worldlythrush, thriftysnipe, decimalswift, solemnlollies, grumpyzebra, dopeytacos},
            style={lines},
          ]{256/worldlythrush, 92/thriftysnipe, 62/decimalswift, 56/solemnlollies, 56/grumpyzebra, 18/dopeytacos}
        \end{tikzpicture}
    }
    \subfigure[RRC 2021 -- Stage 2]{
        \centering
        \begin{tikzpicture}
          \scriptsize
          \begin{axis}[
              ybar stacked,
              bar width=3pt,
              x=4pt,
              ymin=0,
              xtick=data,
              table/col sep=comma,
              legend style={cells={anchor=west}, legend pos=north west},
              reverse legend=true,
              xticklabels from table={rrc2021_stage2_hist.txt}{Date},
              xticklabel style={font=\tiny,rotate=30},
              yticklabel style={font=\tiny},
              ylabel=Number of submissions,
              xtick pos=left,
              ytick pos=left,
            ]
            \addplot [draw=none, fill=thriftysnipe] table
              [y=thriftysnipe, meta=Date, x expr=\coordindex]
              {rrc2021_stage2_hist.txt};
            \addlegendentry{thriftysnipe}
            \addplot [draw=none, fill=dopeytacos] table [y=dopeytacos,
              meta=Date, x expr=\coordindex] {rrc2021_stage2_hist.txt};
            \addlegendentry{dopeytacos}
            \addplot [draw=none, fill=grumpyzebra] table [y=grumpyzebra,
              meta=Date, x expr=\coordindex] {rrc2021_stage2_hist.txt};
            \addlegendentry{grumpyzebra}
            \legend{} 
          \end{axis}
        \end{tikzpicture}
        \begin{tikzpicture}\clip (-3,-3.3) rectangle (4.5,3);
          \scriptsize
          \tikzset{
            lines/.style={draw=none},
          }
          \pie[sum=auto, text=pin, radius=2,
            color={thriftysnipe, dopeytacos, grumpyzebra},
            style={lines},
          ]{37/thriftysnipe, 36/dopeytacos, 6/grumpyzebra}
        \end{tikzpicture}
    }
    \caption{Number of submissions during RRC 2021 over time (left, one bar corresponds to one day) and per team (right).}
    \label{fig:submission_statistics2021}
\end{figure*}

\end{document}